\documentclass[10pt,twocolumn,letterpaper]{article}

\usepackage{wacv}              

\usepackage{graphicx}
\usepackage{amsmath}
\usepackage{amssymb}
\usepackage{booktabs}
\usepackage{multirow}
\usepackage{subcaption}
\usepackage[accsupp]{axessibility}  

\usepackage[pagebackref,breaklinks,colorlinks]{hyperref}

\usepackage[capitalize]{cleveref}
\crefname{section}{Sec.}{Secs.}
\Crefname{section}{Section}{Sections}
\Crefname{table}{Table}{Tables}
\crefname{table}{Tab.}{Tabs.}

\def\wacvPaperID{3} 
\def\confName{WACV}
\def\confYear{2024}

\begin{document}

\title{FRCSyn Challenge at WACV 2024:\\Face Recognition Challenge in the Era of Synthetic Data}

\author{Pietro Melzi$^{1}$
\and 
Ruben Tolosana$^{1}$
\and
Ruben Vera-Rodriguez$^{1}$
\and
Minchul Kim$^{2}$
\and
Christian Rathgeb$^{3}$
\and 
Xiaoming Liu$^{2}$
\and
Ivan DeAndres-Tame$^{1}$
\and
Aythami Morales$^{1}$
\and
Julian Fierrez$^{1}$
\and 
Javier Ortega-Garcia$^{1}$
\and
Weisong Zhao$^{4,5}$
\and 
Xiangyu Zhu$^{6,7}$
\and
Zheyu Yan$^{6}$
\and 
Xiao-Yu Zhang$^{4,5}$
\and
Jinlin Wu$^{8}$
\and
Zhen Lei$^{6,7,8}$
\and 
Suvidha Tripathi$^{9}$
\and 
Mahak Kothari$^{9}$
\and
Md Haider Zama$^{9}$
\and 
Debayan Deb$^{9}$
\and 
Bernardo Biesseck$^{10,11}$
\and
Pedro Vidal$^{10}$
\and
Roger Granada$^{12}$
\and
Guilherme Fickel$^{12}$
\and
Gustavo Führ$^{12}$
\and
David Menotti$^{10}$
\and
Alexander Unnervik$^{13,14}$
\and 
Anjith George$^{13}$
\and
Christophe Ecabert$^{13}$
\and 
Hatef Otroshi Shahreza$^{13,14}$
\and
Parsa Rahimi$^{13,14}$
\and 
S\'{e}bastien Marcel$^{13,15}$
\and 
Ioannis Sarridis$^{16}$
\and
Christos Koutlis$^{16}$
\and 
Georgia Baltsou$^{16}$
\and
Symeon Papadopoulos$^{16}$
\and 
Christos Diou$^{17}$
\and
Nicolò Di Domenico$^{18}$
\and
Guido Borghi$^{18}$
\and 
Lorenzo Pellegrini$^{18}$
\and
Enrique Mas-Candela$^{19}$
\and 
\'{A}ngela S\'{a}nchez-P\'{e}rez$^{19}$
\and
Andrea Atzori$^{20}$
\and 
Fadi Boutros$^{21,22}$
\and
Naser Damer$^{21,22}$
\and
Gianni Fenu$^{20}$
\and 
Mirko Marras$^{20}$
\and
\normalsize{$^{1}$Universidad Autonoma de Madrid, Spain}
\normalsize{$^{2}$Michigan State University, US}
\normalsize{$^{3}$Hochschule Darmstadt, Germany}\\
\normalsize{$^{4}$IIE, CAS, China}
\normalsize{$^{5}$School of Cyber Security, UCAS, China}
\normalsize{$^{6}$MAIS, CASIA, China}\\
\normalsize{$^{7}$School of Artificial Intelligence, UCAS, China}
\normalsize{$^{8}$CAIR, HKISI, CAS, China}
\normalsize{$^{9}$LENS, Inc., USA}\\
\normalsize{$^{10}$Federal University of Paraná, Curitiba, PR, Brazil}
\normalsize{$^{11}$Federal Institute of Mato Grosso, Pontes e Lacerda, Brazil}\\
\normalsize{$^{12}$unico - idTech, Brazil}
\normalsize{$^{13}$Idiap Research Institute, Switzerland}
\normalsize{$^{14}$\'{E}cole Polytechnique F\'{e}d\'{e}rale de Lausanne, Switzerland}\\
\normalsize{$^{15}$Universit\'{e} de Lausanne, Switzerland}
\normalsize{$^{16}$Centre for Research and Technology Hellas, Greece}\\
\normalsize{$^{17}$Harokopio University of Athens, Greece}
\normalsize{$^{18}$University of Bologna, Cesena Campus, Italy}
\normalsize{$^{19}$Facephi, Spain}\\
\normalsize{$^{20}$University of Cagliari, Italy}
\normalsize{$^{21}$Fraunhofer IGD, Germany}
\normalsize{$^{22}$TU Darmstadt, Germany}}
\maketitle

\begin{abstract}
\vspace*{-.4em}
  Despite the widespread adoption of face recognition technology around the world, and its remarkable performance on current benchmarks, there are still several challenges that must be covered in more detail. This paper offers an overview of the Face Recognition Challenge in the Era of Synthetic Data (FRCSyn) organized at WACV 2024. This is the first international challenge aiming to explore the use of synthetic data in face recognition to address existing limitations in the technology. Specifically, the FRCSyn Challenge targets concerns related to data privacy issues, demographic biases, generalization to unseen scenarios, and performance limitations in challenging scenarios, including significant age disparities between enrollment and testing, pose variations, and occlusions. The results achieved in the FRCSyn Challenge, together with the proposed benchmark, contribute significantly to the application of synthetic data to improve face recognition technology.
\end{abstract}

\section{Introduction}
\label{sec:intro}
Facial images represent the most popular data for biometric recognition nowadays, finding extensive applications in surveillance, government offices, and smartphone authentication \cite{minaee2023biometrics}, among others. Numerous studies in the literature have contributed to the development of state-of-the-art (SOTA) Face Recognition (FR) technologies, demonstrating exceptional performance on standard benchmarks \cite{deng2019arcface, kim2022adaface}. The success of these technologies is attributed to the advent of Deep Learning (DL) and the formulation of highly effective loss functions based on margin loss, capable of generating highly discriminative features \cite{wang2021deep}. As a result, FR systems have significantly advanced, achieving astonishing results on well-recognized databases, such as LFW \cite{huang2008labeled}.

\begin{figure*}[t]
    \centering
    \begin{subfigure}{0.46\textwidth}
        \centering
        \includegraphics[width=\textwidth]{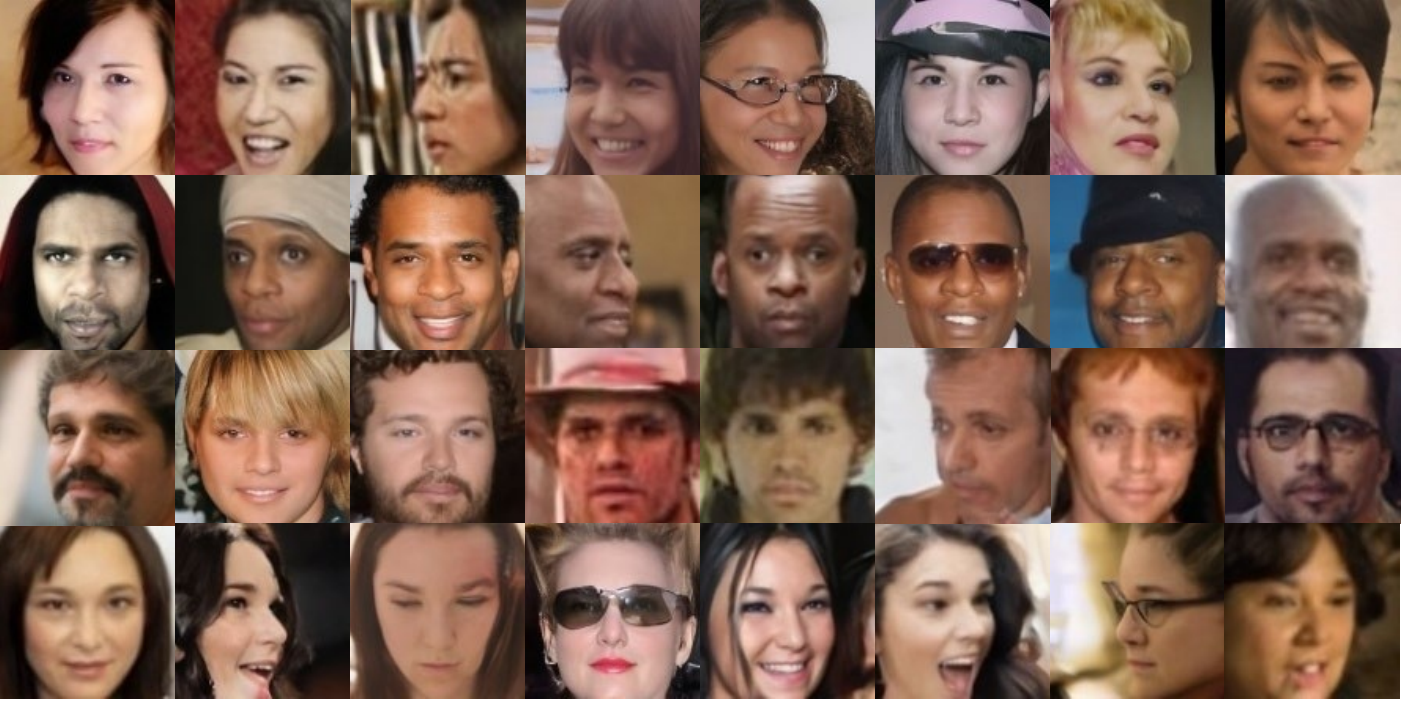}
        \caption{DCFace \cite{kim2023dcface}.}
    \end{subfigure}\hfill
    \begin{subfigure}{0.46\textwidth}
        \centering
        \includegraphics[width=\textwidth]{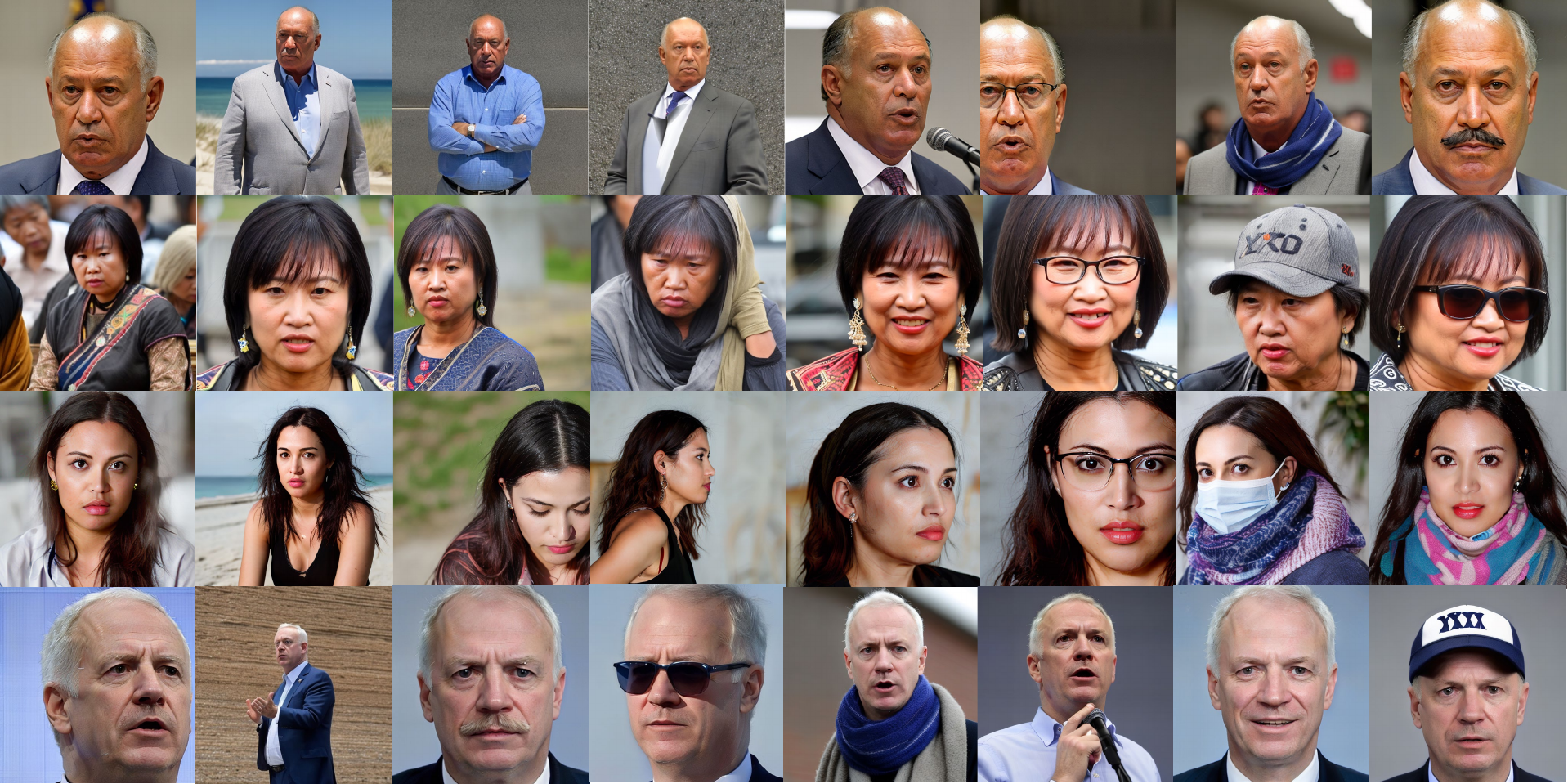}
        \caption{GANDiffFace \cite{melzi2023gandiffface}.}
    \end{subfigure}
        
    \caption{Examples of synthetic identities (one for each row) and intra-class variations for different demographic groups.}
    \label{fig:examples}
\end{figure*}

However, FR still encounters numerous challenges due to factors such as variations in facial images concerning pose, aging, expressions, and occlusions, giving rise to significant issues in the field \cite{adjabi2020past, wanyonyi2022open, minaee2023biometrics}. The application of DL introduces additional concerns, including limited training data, noisy labeling, imbalanced data related to different identities and demographic groups, and low resolution, among other issues \cite{du2022elements}. Deploying FR systems that remain resilient to these challenges and generalize well to unseen conditions is a difficult task. For instance, training data often exhibit significant imbalances across demographic groups \cite{wang2021deep} and may fail to represent the full spectrum of possible occlusions in real-world scenarios \cite{zeng2021survey}. Various limitations associated with established databases and benchmarks are discussed in \cite{ali2021classical}.

In recent years, several approaches have been presented in the literature for the generation of face synthetic content \cite{deng2020disentangled, bae2023digiface, zhang2023iti} for different applications such as FR \cite{kim2023dcface, melzi2023synthetic, boutros2023synthetic} and digital face manipulations, a.k.a. DeepFakes \cite{tolosana2020deepfakes, rathgeb2022handbook, neves2020ganprintr}. These synthetic data offer several advantages over real-world databases. Firstly, synthetic databases provide a promising solution to address privacy concerns associated with real data, often collected from individuals without their knowledge or consent through various online sources \cite{murgia2019s}. Secondly, synthetic face generators have the potential to produce large amounts of data, especially valuable following the discontinuation of established databases due to privacy concerns \cite{Exposing_ai} and the enforcement of regulations like the EU-GDPR, which requires informed consent for collecting and using personal data \cite{voigt2017eu}. Finally, when the synthesis process is controllable, it becomes relatively straightforward to create databases with the desired characteristics (\emph{e.g.,} demographic groups, age, pose, etc.) and their corresponding labels, without additional human efforts. This contrasts with real-world databases, which may not adequately represent diverse demographic groups \cite{morales2020sensitivenets}, among many other aspects.

These advantages have motivated an initial exploration of the application of face synthetic data to current FR systems. Innovative generative frameworks have been introduced to synthesize databases suitable for training FR systems, including Generative Adversarial Networks (GANs) \cite{qiu2021synface, boutros2022sface} and 3D models \cite{bae2023digiface}. While these synthetic databases advance in the field, some have limitations that impact FR systems performance compared to those trained with real data. Specifically, databases synthesized with GANs provide limited representations of intra-class variations \cite{qiu2021synface}, and those synthesized with 3D models lack realism. Recently, Diffusion models have been employed to generate synthetic databases with enhanced intra-class variations, effectively mitigating some limitations observed in prior synthetic databases \cite{kim2023dcface, melzi2023gandiffface}. This is supported by various recent works involving Diffusion models \cite{boutros2023idiff, kansy2023controllable, zhang2023iti}.

To evaluate the effectiveness of novel synthetic databases generated using Diffusion models for training FR systems, this paper analyzes the results achieved in the ``Face Recognition Challenge in the Era of Synthetic Data (FRCSyn)'' organized at WACV 2024\footnote{\url{https://frcsyn.github.io/}}. This challenge is designed to comprehensively analyze the following research questions: 
\begin{enumerate}
    \item Can synthetic data effectively replace real data for training FR systems, and what are the limits of FR technology exclusively trained with synthetic data?
    \item Can the utilization of synthetic data be beneficial in addressing and mitigating the existing limitations within FR technology?
\end{enumerate}
In the proposed FRCSyn Challenge, we have designed specific tasks and sub-tasks to address these questions. In addition, we have released to the participants two novel synthetic databases created using two state-of-the-art Diffusion methods: DCFace \cite{kim2023dcface} and GANDiffFace \cite{melzi2023gandiffface}. These databases have been generated with a particular focus on tackling common challenges in FR, including imbalanced demographic distributions, pose variation, expression diversity, and the presence of occlusion (see Figure \ref{fig:examples}).


The proposed FRCSyn Challenge provides valuable insights for the future of FR and the utilization of synthetic data, with a specific emphasis on quantifying the performance gap between training FR systems with real and synthetic data. In addition, the FRCSyn Challenge proposes standard benchmarks that are easily reproducible for the research community. The reminder of the paper is organized as follows. 
Section \ref{sec:database} provides details about the databases considered in the FRCSyn Challenge. In Section \ref{sec:frcsyn}, we outline the proposed tasks and sub-tasks, the experimental protocol, and metrics used in the challenge. In Section \ref{sec:description}, we provide a description of the top-5 FR systems proposed in the FRCSyn Challenge for each sub-task. Section \ref{sec:results} presents the results achieved in the different tasks and sub-tasks of the challenge. Finally, in Section \ref{sec:conclusion}, we draw the conclusions from the FRCSyn Challenge and highlight potential future research directions in the field.

\section{FRCSyn Challenge: Databases} \label{sec:database}
Table \ref{tab:1} provides details of the public databases considered in the FRCSyn Challenge. Participants were instructed to download all necessary databases for the FRCSyn Challenge upon registration. Permission for redistributing these databases was obtained from the owners.

\begin{table}[]
\centering
\resizebox{0.47\textwidth}{!}{\begin{tabular}{lllcc}
\textbf{Database}     & \textbf{Framework}      & \textbf{Use}   & \textbf{\# Id} & \textbf{\# Img/Id} \\ \hline
DCFace \cite{kim2023dcface} & DCFace & Train & 10K        & 50                                                             \\ 
GANDiffFace \cite{melzi2023gandiffface} & GANDiffFace & Train & 10K        & 50    \\ \hline
CASIA-WebFace \cite{yi2014learning}   & Real-world    & Train     & 10.5K & 47 \\ 
FFHQ \cite{karras2019style}            & Real-world    &  Train    & 70K & 1 \\ \hline
BUPT-BalancedFace \cite{wang2020mitigating}        & Real-world      & Eval  & 24K        & 45                                         \\ 
AgeDB \cite{moschoglou2017agedb}            & Real-world    &  Eval    & 570 & 29 \\ 
CFP-FP \cite{sengupta2016frontal}            & Real-world    &  Eval    & 500 & 14 \\ 
ROF \cite{erakiotan2021recognizing} & Real-world & Eval & 180 & 31 \\ \hline
\end{tabular}}
\caption{Details of the databases considered in the FRCSyn Challenge. Id = Identities, Img = Images.}
\label{tab:1}
\end{table}

\paragraph{Synthetic Databases:}
For the training of the proposed FR systems, we provide access to two synthetic databases generated using recent frameworks based on Diffusion models: 
\begin{itemize}
    \item \textbf{DCFace} \cite{kim2023dcface}. This framework comprises: \emph{i)} a sampling stage for generating synthetic identities $X_{ID}$, and \emph{ii)} a mixing stage for generating images $X_{ID, sty}$ with the same identities $X_{ID}$ from the sampling stage and styles selected from a ``style bank'' of images $X_{sty}$.
    \item \textbf{GANDiffFace} \cite{melzi2023gandiffface}. This framework combines GANs and Diffusion models to generate fully-synthetic FR databases with desired properties such as human face realism, controllable demographic distributions, and realistic intra-class variations.
\end{itemize}

Figure \ref{fig:examples} provides examples of the synthetic face images created using DCFace and GANDiffFace approaches. These synthetic databases represent a diverse range of demographic groups, including variations in ethnicity, gender, and age. The synthesis process considers typical variations in FR, including pose, facial expression, illumination, and occlusions. In the FRCSyn Challenge, synthetic data are exclusively utilized in the training stage, replicating realistic operational scenarios. 

\paragraph{Real Databases:}
For the training of FR systems (depending on the sub-task, please see Section \ref{sec:frcsyn}), participants are allowed to use two real databases: \textit{i)} \textbf{CASIA-WebFace} \cite{yi2014learning}, a database containing $494,414$ face images of $10,575$ real identities collected from the web, and \textit{ii)} \textbf{FFHQ} \cite{karras2019style}, a database designed for face applications, containing $70,000$ high-quality face images with considerable variation in terms of age, ethnicity and image background. These real databases are chosen as they are used to train the generative frameworks of DCFace and GANDiffFace, respectively. This strategy enables a direct comparison between the traditional approach of training FR systems using only real data and the novel approach explored in this challenge, using synthetic data. Despite not being specifically designed for face recognition, the FFHQ database can be considered in the proposed challenge for various purposes, such as training a model for feature extraction and applying domain adaptation, among other possibilities. 

For the final evaluation of the proposed FR systems, we consider four real databases: \textit{i)} \textbf{BUPT-BalancedFace} \cite{wang2020mitigating}, \textit{ii)} \textbf{AgeDB} \cite{moschoglou2017agedb}, \textit{iii)} \textbf{CFP-FP} \cite{sengupta2016frontal}, and \textit{iv)} \textbf{ROF} \cite{erakiotan2021recognizing}. BUPT-BalancedFace \cite{wang2020mitigating} is designed to address performance disparities across different ethnic groups. We relabel it according to the FairFace classifier \cite{karkkainen2021fairface}, which provides labels for ethnicity and gender. We then consider the eight demographic groups obtained from all possible combinations of four ethnic groups (Asian, Black, Indian, and White) and two genders (Female and Male). We recognize that these groups do not comprehensively represent the entire spectrum of real world ethnic diversity. The selection of these categories, while imperfect, is primarily driven by the need to align with the demographic categorizations used in BUPT-BalancedFace \cite{wang2020mitigating} for facilitating easier and more consistent evaluation. The other three databases, \textit{i.e.,} AgeDB \cite{moschoglou2017agedb}, CFP-FP \cite{sengupta2016frontal}, and ROF \cite{erakiotan2021recognizing}, are real-world databases widely employed to benchmark FR systems in terms of age variations, pose variations, and presence of occlusions. It is important to highlight that, as different real databases are considered for training and evaluation, we also intend to analyse the generalization ability of the proposed FR systems.

\section{FRCSyn Challenge: Setup} \label{sec:frcsyn}

\subsection{Tasks}
The FRCSyn Challenge has been hosted on Codalab\footnote{\url{https://codalab.lisn.upsaclay.fr/competitions/15485}}, an open-source framework for running scientific competitions and benchmarks. It aims to explore the application of synthetic data into the training of FR systems, with a specific focus on addressing two critical aspects in current FR technology: \emph{i)} mitigating demographic bias, and \emph{ii)} enhancing overall performance under challenging conditions that include variations in age and pose, the presence of occlusions, and diverse demographic groups. To investigate these two areas, the FRCSyn Challenge considers two distinct tasks, each comprising two sub-tasks. Sub-tasks have been designed to consider different approaches for training FR systems: \textit{i)} utilizing solely synthetic data, and \textit{ii)} involving a combination of real and synthetic data. Consequently, the FRCSyn Challenge comprises a total of four sub-tasks. A summary is provided in Table \ref{tab:2}. For each sub-task, we specify the databases allowed for training FR systems. Nevertheless, participants have the flexibility to decide whether and how to utilize each database in the training process. 

\begin{table}[]
    \centering
    \resizebox{0.47\textwidth}{!}{\begin{tabular}{|l|} \hline
         \textbf{Task 1:} synthetic data for \textbf{demographic bias mitigation} \\
         \quad Baseline: training only with CASIA-WebFace \cite{yi2014learning} and FFHQ \cite{karras2019style}; \\
         \quad Metrics: accuracy (for each demographic group);\\
         \quad Ranking: average vs SD of accuracy, see Section \ref{sec:metrics} for more details. \\ \hline
         \textbf{\textcolor{blue}{Sub-Task 1.1:}} training exclusively with \textbf{synthetic} databases \\
         \quad Train: DCFace \cite{kim2023dcface} and GANDiffFace \cite{melzi2023gandiffface}; \\
         \quad Eval: BUPT-BalancedFace \cite{wang2020mitigating}. \\ \hline
         \textbf{\textcolor{blue}{Sub-Task 1.2:}} training with \textbf{real and synthetic} databases \\ 
         \quad Train: CASIA-WebFace, FFHQ, DCFace, and GANDiffFace; \\
         \quad Eval: BUPT-BalancedFace. \\ \hline \hline
         \textbf{Task 2:} synthetic data for \textbf{overall performance improvement} \\
         \quad Baseline: training only with CASIA-WebFace and FFHQ; \\
         \quad Metrics: accuracy (for each evaluation database);\\
         \quad Ranking: average accuracy. \\ \hline
         \textbf{\textcolor{blue}{Sub-Task 2.1:}} training exclusively with \textbf{synthetic} databases \\ 
         \quad Train: DCFace and GANDiffFace; \\
         \quad Eval: BUPT-BalancedFace, AgeDB \cite{moschoglou2017agedb}, CFP-FP \cite{sengupta2016frontal}, and ROF \cite{erakiotan2021recognizing}. \\ \hline
         \textbf{\textcolor{blue}{Sub-Task 2.2:}} training with \textbf{real and synthetic} databases \\
         \quad Train: CASIA-WebFace, FFHQ, DCFace, and GANDiffFace; \\
         \quad Eval: BUPT-BalancedFace, AgeDB, CFP-FP, and ROF. \\ \hline
    \end{tabular}}
    \caption{Tasks and sub-tasks proposed in FRCSyn Challenge with their respective metrics and databases. SD = Standard Deviation. }
    \label{tab:2}
\end{table}

\paragraph{Task 1:}
The first proposed task explores the use of synthetic data to address demographic biases in FR systems. To evaluate the proposed systems, we create lists of mated and non-mated comparisons derived from individuals in the BUPT-BalancedFace database \cite{wang2020mitigating}. We consider the eight demographic groups described in Section \ref{sec:database}, obtained from the combination of four ethnic groups with two genders. For non-mated comparisons, we exclusively focus on pairs of individuals belonging to the same demographic group, as these are more relevant than non-mated comparisons between individuals of different demographic groups. 

\paragraph{Task 2:}
The second proposed task explores the application of synthetic data to enhance overall performance in FR under challenging conditions. To assess the proposed systems, we use lists of mated and non-mated comparisons derived from individuals included in the four databases indicated in Section \ref{sec:database}, namely BUPT-BalancedFace \cite{wang2020mitigating}, AgeDB \cite{moschoglou2017agedb}, CFP-FP \cite{sengupta2016frontal}, and ROF \cite{erakiotan2021recognizing}. Each database allows the evaluation of specific challenging conditions for FR, including diverse demographic groups, aging, pose variations, and presence of occlusions. 

\subsection{Experimental protocol}

\paragraph{Training:}

The four sub-tasks proposed in the FRCSyn Challenge are mutually independent. This means that participants have the freedom to participate in any number of sub-tasks of their choice. For each selected sub-task, participants are expected to propose a FR system and train it twice: \textit{i)} using authorized real databases only, \textit{i.e.,} CASIA-WebFace \cite{yi2014learning} and FFHQ \cite{karras2019style}, and \textit{ii)} in accordance with the specific requirements of the chosen sub-task, as summarized in Table \ref{tab:2}. According to this protocol, participants provide both the \textit{baseline system} and the \textit{proposed system} for the specific sub-task. The baseline system plays a critical role in evaluating the impact of synthetic data on training and serves as a reference point for comparing against the conventional practice of training solely with real databases. To maintain consistency, the baseline FR system, trained exclusively with real data, and the proposed FR system, trained according to the specifications of the selected sub-task, must have the same architecture.

\paragraph{Evaluation:}
In each sub-task, participants are provided with comparison files containing both mated and non-mated comparisons, which are used to evaluate the performance of their proposed FR system. In Task 1 there is a single comparison file containing balanced comparisons of different demographic groups, while in Task 2 there are four comparison files, one for each real database considered. The evaluation process occurs twice for each sub-task to assess: \textit{i)} the baseline system trained exclusively with real databases, and \textit{ii)} the proposed system trained in accordance with the sub-task specifications. For the evaluation of each sub-task, participants must submit through Codalab platform two files per database (one for the baseline and one for the proposed system), including the score and the binary decision (mated/non-mated) for each comparison listed in the comparison files. The organizers retain the right to disqualify participants to uphold the integrity of the evaluation process if anomalous results are detected or if participants fail to adhere to the challenge's rules.

\paragraph{Restrictions:}
Participants have the freedom to choose the FR system for each task, provided that the system's number of Floating Point Operations Per Second (FLOPs) does not exceed 25 GFLOPs. This threshold has been established to facilitate the exploration of innovative architectures and encourage the use of diverse models while preventing the dominance of excessively large models. Participants are also free to utilize their preferred training modality, with the requirement that only the specified databases are used for training. This means that no additional databases can be employed during the training phase, such as to establish verification thresholds. Generative models cannot be utilized to generate supplementary data. Participants are allowed to use non-face databases for pre-training purposes and employ traditional data augmentation techniques using the authorized training databases.

\subsection{Metrics} \label{sec:metrics}
We evaluate FR systems using a protocol based on lists of mated and non-mated comparisons for each sub-task and database. From the binary decisions provided by participants, we calculate verification accuracy. This approach is straightforward and allows participants to choose the preferred threshold for their systems. Additionally, we calculate the gap to real (GAP) \cite{kim2023dcface} as follows: $\text{GAP} = \left( \text{REAL} - \text{SYN} \right)/\text{SYN}$, with $\text{REAL}$ representing the verification accuracy of the baseline system and $\text{SYN}$ the verification accuracy of the proposed system, trained with synthetic (or real + synthetic) data.
Other metrics such as False Non-Match Rate (FNMR) at different operational points, which are very popular for the analysis of FR systems in real-world applications, can be computed from the scores provided by participants. Comprehensive evaluations of the proposed systems will be conducted in subsequent studies, including FNMRs and metrics for each demographic group and database used for evaluation. Next, we explain how participants are ranked in the different tasks.

\paragraph{Task 1:}
To rank participants and determine the winners of Sub-Tasks 1.1 and 1.2, we closely examine the trade-off between the average (AVG) and standard deviation (SD) of the verification accuracy across the eight demographic groups defined in Section \ref{sec:database}. We define the trade-off metric (TO) as follows: $\text{TO} = \text{AVG} - \text{SD}$. This metric corresponds to plotting the average accuracy on the x-axis and the standard deviation on the y-axis in 2D space. We draw multiple 45-degree parallel lines to find the winning team whose performance falls to the far right side of these lines. With this proposed metric, we reward FR systems that achieve good levels of performance and fairness simultaneously, unlike common benchmarks based only on recognition performance. The standard deviation of verification accuracy across demographic groups is a common metric for assessing bias and should be reported by any work addressing demographic bias mitigation. 

\paragraph{Task 2:}
To rank participants and determine the winners of Sub-Tasks 2.1 and 2.2, we consider the average verification accuracy across the four databases used for evaluation, described in Section \ref{sec:database}. This approach allows us to evaluate four challenging aspects of FR simultaneously: \textit{i)} pose variations, \textit{ii)} aging, \textit{iii)} presence of occlusions, and \textit{iv)} diverse demographic groups, providing a comprehensive evaluation of FR systems in real operational scenarios.

\begin{table}[]
\centering
\resizebox{0.47\textwidth}{!}{
\begin{tabular}{lclc}
 \textbf{Team}                                                    & \textbf{Affiliations} & \textbf{Country} & \textbf{Sub-Tasks} \\ \hline
CBSR                                                             & 4-8                   & China            & 1.2 - 2.2             \\
LENS                                                             & 9                     & USA              & all             \\
BOVIFOCR-UFPR                                                    & 10-12                 & Brazil           & all          \\
Idiap                                                            & 13-15              & Switzerland      & all         \\
MeVer                                                        & 16,17                 & Greece           & all       \\
BioLab                                                           & 18                    & Italy            & 2.1          \\
Aphi                                                             & 19                    & Spain            & 1.1 - 2.1          \\
\begin{tabular}[c]{@{}l@{}}UNICA-FRAUN-\\ HOFER IGD\end{tabular} & 20-22              & \begin{tabular}[c]{@{}l@{}}Italy, \\ Germany\end{tabular}   & 1.2 - 2.2
\\ \hline
\end{tabular}}
\caption{Description of the top-5 best teams ordered by the affiliation number. The numbers reported in the column `affiliations' refer to the ones provided in the title page.}
\label{tab:best_teams}
\end{table}



\section{FRCSyn Challenge: Description of Systems} \label{sec:description}
The FRCSyn Challenge received significant interest, with $67$ international teams correctly registered, comprising research groups from both industry and academia. These teams work in various domains, including FR, generative AI, and other aspects of computer vision, such as demographic fairness and domain adaptation. Finally, we received submissions from $15$ teams, receiving all sub-tasks high attention. The submitting teams are geographically distributed, with six teams from Europe, five teams from Asia, and four teams from America. Table \ref{tab:best_teams} provides a general overview of the top-5 best teams, including the sub-tasks they participated. Next, we describe briefly the FR systems proposed for each team. 

\paragraph{CBSR (Sub-Tasks 1.2 and 2.2):}
They first trained a recognition model using CASIA-WebFace \cite{yi2014learning}. They extracted features for images in FFHQ \cite{karras2019style} and clustered them using the DBSCAN \cite{ester1996density} for pseudo labels. Then, they removed the samples in FFHQ that are similar to CASIA-WebFace with a cosine similarity threshold of $0.6$ and merged the two to train a new model $F$. They utilized $F$ to de-overlap DCFace \cite{kim2023dcface} and GANDiffFace \cite{melzi2023gandiffface} from CASIA-WebFace and FFHQ. Subsequently, they conducted the intra-class clustering for all databases using DBSCAN (similarity threshold of $0.3$) and removed the samples that were separate from the class center. They merged the cleansed databases and trained IResNet-100 with mask and sunglasses augmentation and AdaFace loss \cite{kim2022adaface}. They trained two recognition models using occlusion augmentation with 10\% and 30\% probability, respectively. They finally submitted the average similarity prediction of the two models. The threshold was determined by the 10-fold optimal threshold in the validation set.

They constructed different validation sets for different evaluation tasks. For AgeDB \cite{moschoglou2017agedb}, they randomly sampled pairs from the training databases. For CFP-FP \cite{sengupta2016frontal}, they added randomly positioned vertical bar masks to the images to simulate the self-occlusion due to pose. For ROF \cite{erakiotan2021recognizing}, they detected face landmarks \cite{wang2021facex} and added mask and sunglasses to images. For BUPT-BalancedFace \cite{wang2020mitigating}, they randomly sampled pairs from DCFace with GANDiffFace because they have balanced demographic groups. All validation sets consisted of $12,000$ image pairs containing $6,000$ positive pairs and $6,000$ negative pairs.
Code available\footnote{\url{https://github.com/zws98/wacv_frcsyn}}.

\paragraph{LENS (All sub-tasks):}
For sub-tasks using only synthetic data (\textit{i.e.,} 1.1 and 2.1), they observed that since the evaluation data are real databases, they needed an approach that makes the architecture robust to domain shifts between synthetic training data and real test data. For the same, they utilized the augmentations and AdaFace loss introduced in \cite{kim2022adaface}. The augmentations like Crop, Photometric jittering, and Low-res scaling from \cite{kim2022adaface} helped to create more robust images similar to the real domain, effectively improving performance. They further enhanced the features by using an ensemble of two models, with different styles of augmenting databases like randomly selecting four from set of  Identity, Spatial transformations, Brightness, Color, Contrast, Sharpness, Posterize, Solarize, AutoContrast, Equalize, Grayscale, ResizedCrop augmentations in each iteration, inspired from \cite{boutros2023idiff}. The features of the two models were then combined to create a feature set of length $1024$. The same method was repeated for Sub-Tasks 1.2 and 2.2. 

After cropping and alignment, they divided their total data in the ratio $80:20$ for training and validation, respectively. For training the baseline model and Sub-Tasks 1.2 and 2.2, they utilized CASIA-WebFace \cite{yi2014learning} for the real database and skipped FFHQ \cite{karras2019style}. They adopted the architecture of ResNet-50 \cite{he2016deep} (R50) backbone for all the sub-tasks for its lesser number of parameters and suitability when the size of the databases is not huge. They used AdaFace loss from \cite{kim2022adaface}. 

\paragraph{BOVIFOCR-UFPR (All sub-tasks):}
Inspired by Zhang \textit{et al.} \cite{zhang2018mitigating}, they reduced bias in Sub-Task 1.1 by creating a multi-task collaborative model composed of two backbones $B(x)$ and $R(e)$, which produced the embeddings $e \in R^{512}$ and $g \in R^{256}$, respectively. This schema forced $B(x)$ to learn less biased features across different ethnic groups. ResNet100 and ResNet18 \cite{he2016deep} architectures were used as $B(x)$ and $R(e)$. Each training sample $x_i$ contained two labels $y_i$ (to compute the subject loss $L_S$ \cite{deng2019arcface}), and $w_i$, (to compute the ethnic group loss $L_E$ \cite{deng2019arcface}). Their total loss was $L_T = \lambda_S L_S + \lambda_E L_E$. In Sub-Task 2.1 they employed ArcFace \cite{deng2019arcface} as their loss function and Resnet100 \cite{he2016deep} as the backbone, which is one of the top-performing models for deep FR \cite{deng2021masked}. They trained the network using the InsightFace library for $26$ epochs. The images used for training were augmented using Random Flip with a probability of $0.5$. They used DCFace \cite{kim2023dcface} as the training database in this sub-task, which provided the most accurate feature vectors on the validation set.

\paragraph{Idiap (All sub-tasks):}
The primary strategy for all tasks and sub-tasks was the fusion of features from two models, chosen for its potential to enhance accuracy and reduce bias. These models compute a mean feature vector via a feature fusion approach and undergo independent training to maximize the differences between them, to improve fusion results.
For preprocessing, RetinaFace \cite{deng2020retinaface} was used to detect facial landmarks across all evaluation sets, and a similarity transform aligned five key facial points to a standard template before cropping and resizing images to $112 \times 112$ pixels, with pixel values normalized between $\left[-1, 1\right]$. 

The models were based on iResNet-50 and iResNet-101 architectures. Training utilized specific databases for each track, with the iResNet-101 leveraging CosFace loss \cite{wang2018cosface} and the iResNet-50 using AdaFace loss \cite{kim2022adaface}. Training ran for approximately $60,000$ batches of size $256$, with learning rate adjustments at set intervals. Training data underwent further preprocessing, including random cropping and augmentations in resolution, brightness, contrast, and saturation.
The final model checkpoint was taken after the last training step. A subset of the training data was used to determine the optimal threshold for maximizing verification accuracy, using a 10-fold cross-validation approach based on a random selection of identities and comparison pairs.

\paragraph{MeVer (All sub-tasks):}
Their proposed system utilized the sub-center ArcFace loss \cite{deng2020sub} to mitigate noise, which occurs in synthetic training data \cite{cheng2020learning}. Comprising three CNNs, the proposed system adapted various margins within the ArcFace loss \cite{deng2019arcface}, aligning with relevant literature, indicating different demographic groups require different margin considerations \cite{wang2021meta}. Final embeddings were obtained by combining the outputs of three ResNet-50 \cite{he2016deep} models each trained with $4$, $5$, and $5$ subcenters and margins of $0.45$, $0.47$, and $0.50$. Prediction involved computing the Euclidean distance between feature vectors, utilizing thresholds of $1.5$ and $1.35$ for tasks involving synthetic-only and mixed synthetic-real training data, respectively. The training procedure involved a batch size of $256$, an initial learning rate of $0.1$ that decayed by a factor of $10$ at steps $75$k, $127.5$k, and $165$k over $180$k total training steps. Optimizing with stochastic gradient descent (SGD), momentum was set at $0.9$, and weight decay at $0.0005$. Data preprocessing involved an MTCNN \cite{zhang2016joint}, resizing all data to $112 \times 112$, and employing color jittering and random horizontal flip augmentations. Task-wise, both synthetic databases were utilized, while the CASIA-WebFace database was specific to Sub-Tasks 1.2 and 2.2. Validation included $800$ synthetic identities and $1,000$ identities from CASIA-WebFace for the tasks involving synthetic-only and mixed synthetic-real databases, respectively. Code available\footnote{\url{https://github.com/gsarridis/fair-face-verification-with-synthetic-data}}.

\paragraph{BioLab (Sub-Task 2.1):}
The model selected for the Sub-Task 2.1 is a customized ResNet-101 \cite{deng2019arcface, he2016deep}, which had been trained using the margin-based AdaFace loss \cite{kim2022adaface}, whose advantage is its resilience when training data contain low-quality images with unrecognizable faces. According to their assumption, this ensured that the model's performance remained unaffected when exposed to GAN-related visual glitches and artifacts. Their baseline model was trained employing the CASIA-WebFace database \cite{yi2014learning}. Differently, the proposed model employed both DCFace \cite{kim2023dcface} and GANDiffFace \cite{melzi2023gandiffface}. In both cases they built the validation set by generating couples from the first classes of the training sets, which were excluded from training.
They applied data augmentation on the training set. Following \cite{kim2022adaface}, the pipeline consisted of random horizontal flips, random crop-and-resize, and random color jittering on saturation and value channels. Each transformation had a probability of $20\%$ of being applied. Finally, the model was optimized with cross entropy loss and SGD with an initial learning rate of $0.05$. Learning rate scheduling was employed to improve training stability.
For face verification, the dissimilarity between embeddings was measured employing the cosine distance. Its threshold was computed to maximize the accuracy on the validation set (\textit{i.e.,} using a non-overlapping partition of the training databases), following the same idea described in the LFW protocol \cite{huang2008labeled}. Code available\footnote{\url{https://github.com/ndido98/frcsyn}}.

\paragraph{Aphi (Sub-Tasks 1.1 and 2.1):}
In their approach, they used an EfficientNetV2-S \cite{tan2021efficientnetv2} architecture to produce a $512$-D deep embedding trained with ArcFace \cite{deng2019arcface} loss function. They modified the backbone network by reducing the first layer's stride from $2$ to $1$ to enhance the preservation of spatial features. The output of the backbone network was projected with a $1 \times 1$ convolutional layer and normalized with batch normalization. These features were flattened and fed into a fully connected layer which produces the deep embedding. The weights of the model were optimized through the SGD algorithm with a momentum of $0.9$ and a weight decay of $1e^{-4}$ during $20$ epochs and a learning rate starting at $0.1$ and decayed through a polynomial scheduler.
The model was trained with the images aligned using a proprietary algorithm, resized to $112 \times 112$, and normalized in the range of $-1$ to $1$. To prevent overfitting, they applied data augmentation techniques during training, including Gaussian Blur, Random Scale, Hue-Saturation adjustments, and Horizontal Flip transformations as well as dropout with a rate of $0.2$ before the deep embedding projection.
To train the baseline model, they made use of CASIA-WebFace \cite{yi2014learning} and for their proposed model, they employed the synthetic database DCFace \cite{kim2023dcface}.

\paragraph{UNICA-FRAUNHOFER IGD (Sub-Tasks 1.2 and 2.2):}
The presented solution utilized ResNet100 \cite{he2016deep} as network architecture as it is one of the most widely used architectures in state-of-the-art FR approaches \cite{boutros2022elasticface}. Training and validation images were aligned and cropped to $112 \times 112$ using five-points landmarks extracted with MTCNN. The network's outputs were $512$-D feature representations.
The presented solution, submitted to Sub-Tasks 1.2 and 2.2, relies on training the ResNet100 network with CosFace as a loss function with a margin penalty value of $0.35$ and a scale parameter of $64$ \cite{wang2018cosface}. The model was trained for $40$ epochs with a batch size of $512$ and an initial learning rate of $0.1$. The learning rate was divided by $10$ after $10$, $22$, $30$, and $40$ epochs. During the training phase the training databases, CASIA-WebFace \cite{yi2014learning} and DCFace \cite{kim2023dcface}, provided by the competition organizers, were merged into one database with a total number of $20.572$ identities. During the training phase, an extensive set of data augmentation operations based on RandAugment \cite{boutros2023unsupervised, cubuk2020randaugment} was applied only to the synthetic samples. The real samples were only augmented with horizontal flipping. Code available\footnote{\url{https://github.com/atzoriandrea/FRCSyn}}.

\begin{table}
\footnotesize
\centering
\resizebox{0.47\textwidth}{!}{
\begin{tabular}{clcccc}
\multicolumn{6}{c}{\textbf{Sub-Task 1.1 (Bias Mitigation): Synthetic Data}} \\ \hline
\textbf{Pos.} & \textbf{Team} & \textbf{TO [\%]} & \textbf{AVG [\%]} & \textbf{SD [\%]} & \textbf{GAP [\%]} \\ \hline 
\textbf{1} & \textbf{LENS} & \textbf{92.25} & \textbf{93.54} & \textbf{1.28}  &    \textbf{-0.74}                   \\ 
2 & Idiap         & 91.88 & 93.41 & 1.53 & -3.80                                  \\ 
3 & BOVIFOCR & 90.51 & 92.35 & 1.84 & 4.23                                  \\ 
4 & MeVer     & 87.51 & 89.62 & 2.11 & 5.68                                  \\ 
5 & Aphi          & 82.24 & 86.01 & 3.77 & 0.84                                  \\ \hline
\end{tabular}}
\bigskip

\resizebox{0.47\textwidth}{!}{
\begin{tabular}{clcccc}
\multicolumn{6}{c}{\textbf{Sub-Task 1.2 (Bias Mitigation): Synthetic + Real Data}} \\ \hline
\textbf{Pos.} & \textbf{Team} & \textbf{TO [\%]} & \textbf{AVG [\%]} & \textbf{SD [\%]} & \textbf{GAP [\%]} \\ \hline 
\textbf{1} & \textbf{CBSR} & \textbf{95.25}  & \textbf{96.45} & \textbf{1.20}  & \textbf{-2.10}                     \\ 
2 & LENS         & 95.24      & 96.35 & 1.11     & -5.67                       \\ 
3 & MeVer & 93.87                       & 95.44 & 1.56   & -0.78        \\ 
4 & BOVIFOCR     & 93.15                             & 95.04 & 1.89  & 1.28   \\ 
5 & UNICA         & 91.03                      & 94.06 & 3.03   & -10.62        \\ \hline
\end{tabular}}
\bigskip

\begin{tabular}{clcc}
\multicolumn{4}{c}{\textbf{Sub-Task 2.1 (Overall Improvement): Synthetic Data}} \\ \hline
\textbf{Pos.} & \textbf{Team} & \textbf{AVG [\%]} & \textbf{GAP [\%]} \\ \hline 
\textbf{1} & \textbf{BOVIFOCR} & \textbf{90.50} & \textbf{2.66}                        \\ 
2 & LENS         & 88.18         & 3.75                         \\ 
3 & Idiap & 86.39              & 6.39                    \\ 
4 & BioLab     & 83.93               & 6.88                   \\ 
5 & MeVer          & 83.45             & 3.20                     \\ \hline
\end{tabular}
\bigskip

\begin{tabular}{clcc}
\multicolumn{4}{c}{\textbf{Sub-Task 2.2 (Overall Improvement): Synthetic + Real Data}} \\ \hline
\textbf{Pos.} & \textbf{Team} & \textbf{AVG [\%]} & \textbf{GAP [\%]} \\ \hline 
\textbf{1} & \textbf{CBSR} & \textbf{94.95}            & \textbf{-3.69}             \\ 
2 & LENS         & 92.40                 & -1.63                \\ 
3 & Idiap & 91.74                 & 0.00                  \\ 
4 & BOVIFOCR     & 91.34              & 1.77                       \\ 
5 & MeVer          & 87.60               & -1.57                      \\ \hline
\end{tabular}

\caption{Ranking for the four sub-tasks, according to the metrics described in Section \ref{sec:metrics}. TO = Trade-Off, AVG = Average accuracy, SD = Standard Deviation of accuracy, GAP = Gap to Real.}
\label{tab:results}
\end{table}

\section{FRCSyn Challenge: Results} \label{sec:results}
Table \ref{tab:results} presents the rankings for the different sub-tasks considered in the FRCSyn Challenge. In general, the rankings for Sub-Tasks 1.1 and 1.2 (bias mitigation), corresponding to the descending order of TO, closely align with the ascending order of SD (\textit{i.e.,} from less to more biased FR systems). Notably, in Sub-Task 1.1, the top two classified teams, LENS (92.25\% TO) and Idiap (91.88\% TO), exhibit negative GAP values (-0.74\% and -3.80\%, respectively), indicating higher accuracy when training the FR system with synthetic data compared to real data. These results highlight the potential of DCFace \cite{kim2023dcface} and GANDiffFace \cite{melzi2023gandiffface} synthetic data to reduce bias in current FR technology. The inclusion of real data in the training process (\textit{i.e.,} Sub-Task 1.2) results in general in a simultaneous increase in AVG and reduction in SD, being the CBSR team the winner with a 95.25\% TO (\textit{i.e.,} 3\% TO general improvement between Sub-Tasks 1.1 and 1.2). In addition, and as it happens in Sub-Task 1.1, we can observe in Sub-Task 1.2 negative GAP values for the top teams (\textit{e.g.,} -2.10\% and -5.67\% for the CBSR and LENS teams, respectively), evidencing that the combination of synthetic and real data (proposed system) outperforms FR systems trained only with real data (baseline system).

For Task 2, it is evident that the average accuracy across databases in Sub-Tasks 2.1 and 2.2 is lower than the accuracy achieved for BUPT-BalancedFace \cite{wang2020mitigating} in Sub-Tasks 1.1 and 1.2, emphasizing the additional challenges introduced by the other real databases considered for evaluation. Also, although good results are achieved in Sub-Task 2.1 when training only with synthetic data (90.50\% AVG for BOVIFOCR-UFPR), the positive GAP values provided by the top-5 teams indicate that synthetic data alone currently struggles to completely replace real data for training FR systems in challenging conditions. Nevertheless, the negative GAP values provided by the top-2 teams in Sub-Task 2.2 also suggest that synthetic data combining with real data can mitigate existing limitations within FR technology.

Finally, analyzing the contributions of all the eight top teams, a notable trend emerges, showing the prevalence of well-established methodologies. ResNet backbones \cite{he2016deep} were chosen by seven teams, except for Aphi, which opted for EfficientNet \cite{tan2021efficientnetv2}. The AdaFace \cite{kim2022adaface} and ArcFace \cite{deng2019arcface} loss functions were widely used, featuring in the approaches of CBSR, LENS, Idiap, and BioLab for the former, and BOVIFOCR-UFPR, MeVer, and Aphi for the latter. Idiap and UNICA-FRAUNHOFER IGD also considered the CosFace loss function \cite{wang2018cosface}. Most of the teams integrated multiple networks into their proposed architectures for different objectives, \textit{e.g.}, CBSR and LENS trained different networks with distinct augmentation techniques, while BOVIFOCR-UFPR and Idiap combined different loss functions.
Some teams also addressed the challenges of domain shift between synthetic and real data, \textit{e.g.}, LENS proposed solutions robust to domain shifts with consistent data augmentation, while CBSR implemented a range of strategies, including advanced data augmentation, identity clustering, and distinct thresholds for different databases. Notably, CBSR utilized all available databases for training, including FFHQ \cite{karras2019style}, unlike other teams. Excluding BOVIFOCR-UFPR, Aphi, and UNICA-FRAUNHOFER IGD, which exclusively used DCFace \cite{kim2023dcface}, the majority of teams employed both DCFace \cite{kim2023dcface} and GANDiffFace \cite{melzi2023gandiffface}, demonstrating the suitability of both generative frameworks.

\section{Conclusion} \label{sec:conclusion}
The Face Recognition Challenge in the Era of Synthetic Data (FRCSyn) has provided a comprehensive analysis for the application of synthetic data to FR, addressing current limitations in the field. Within this challenge numerous approaches from different research groups have been proposed. These approaches can be compared across a variety of sub-tasks, with many being reproducible thanks to the materials made available by the participating teams.
Future works will be oriented to a more detailed analysed of the results, including additional metrics and graphical representations. Furthermore, we are considering transforming the CodaLab platform into an ongoing competition, where new tasks and sub-tasks might be introduced. 

\section*{Acknowledgements}
{\footnotesize Special thanks to Mei Wang and Stylianos Moschoglou for authorizing the distribution of their databases.
This study has received funding from the European Union’s Horizon 2020 TReSPAsS-ETN (No 860813) and is supported by INTER-ACTION (PID2021- 126521OB-I00 MICINN/FEDER) and R\&D Agreement DGGC/UAM/FUAM for Biometrics and Cybersecurity.
It is also supported by the German Federal Ministry of Education and Research and the Hessian Ministry of Higher Education, Research, Science and the Arts within their joint support of the National Research Center for Applied Cybersecurity ATHENE.
BioLab acknowledge Andrea Pilzer from the NVIDIA AI Technology Center, EMEA, for his support. 
MeVer was supported by the EU Horizon Europe project MAMMOth (Grant Agreement 101070285).}

{\small
\bibliographystyle{ieee_fullname}
\bibliography{egbib}
}

\end{document}